\documentclass[runningheads]{llncs}
\usepackage[T1]{fontenc}
\usepackage{graphicx,verbatim}
\usepackage{amsmath}
\usepackage{amssymb}
\usepackage{algorithm}
\usepackage{algpseudocode}
\usepackage{booktabs}
\usepackage{multirow}
\usepackage{graphicx}
\usepackage{subcaption}
\usepackage{xcolor}

\begin{document}

\title{RPG-SAM: Reliability-Weighted Prototypes and Geometric Adaptive Threshold Selection for Training-Free One-Shot Polyp Segmentation}
\titlerunning{RPG-SAM}

%
\author{Weikun Lin\inst{1}\thanks{Equal contribution.} \and Yunhao Bai\inst{1}\textsuperscript{\thefootnote} \and Yan Wang\inst{1}}
\authorrunning{W.~Lin, Y.~Bai et al.}
\institute{\textsuperscript{1}Shanghai Key Laboratory of Multidimensional Information Processing, \\ East China Normal University \\
{\tt 10232140464@stu.ecnu.edu.cn}}


  
\maketitle              

\begin{abstract}
Training-free one-shot segmentation offers a scalable alternative to expert annotations where knowledge is often transferred from support images and foundation models. But existing methods often treat all pixels in support images and query response intensities models in a homogeneous way. They ignore the regional heterogeity in support images and response heterogeity in query.
To resolve this, we propose RPG-SAM, a framework that systematically tackles these heterogeneity gaps. Specifically, to address regional heterogeneity, we introduce Reliability-Weighted Prototype Mining (RWPM) to prioritize high-fidelity support features while utilizing background anchors as contrastive references for noise suppression. To address response heterogeneity, we develop Geometric Adaptive Selection (GAS) to dynamically recalibrate binarization thresholds by evaluating the morphological consensus of candidates. Finally, an iterative refinement loop method is designed to  polishes anatomical boundaries. By accounting for multi-layered information heterogeneity, RPG-SAM achieves a 5.56\% mIoU improvement on the Kvasir dataset. Code will be released.
\end{abstract}

\begin{figure}
\centering
\includegraphics[width=0.9\textwidth]{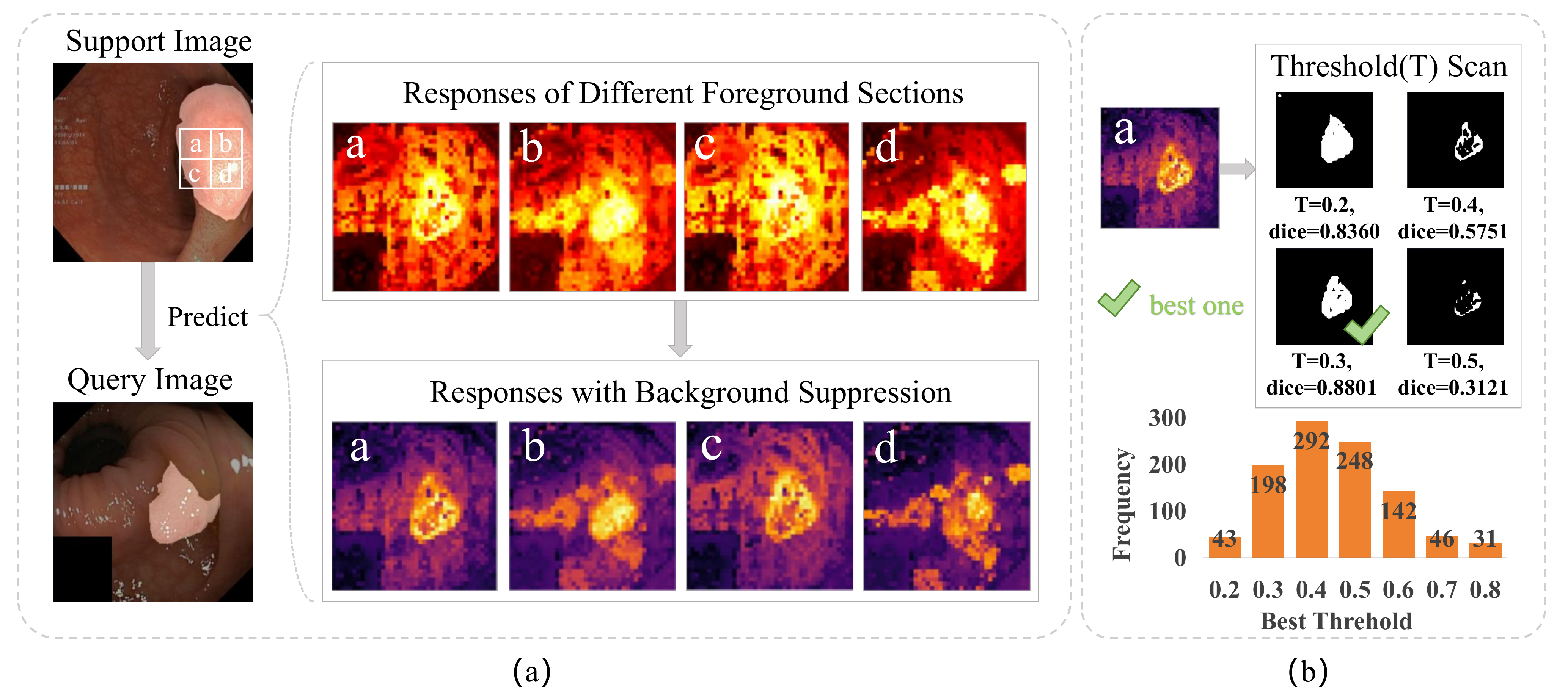}
\caption{Motivation of RPG-SAM. (a) Regional and Contextual Heterogeneity. Top: The varying regional utility of the support foreground means that degraded areas (e.g., reflection $d$) trigger false-positive noise in query images. Bottom: Treating the contextual background as a distinct information layer provides negative anchors to effectively suppress heatmap activations. (b) Intensity Heterogeneity. The histogram of optimal binarization thresholds across the Kvasir dataset \cite{datasetKvasir} highlights the stochastic response intensities of different query scenarios, exposing the limitations of fixed-threshold (homogeneous) sampling rules. } \label{fig:motivation}
\end{figure}

\section{Introduction}
Polyp detection is a key task in early colorectal cancer (CRC) screening\cite{zauber2012colonoscopic,sung2021global,jain2023coinnet}. Although supervised models have achieved high accuracy\cite{fan2020pranet,huang2021hardnet,rahman2023medical}, their dependence on large-scale pixel-level annotations limits clinical scalability. Recently, training-free one-shot segmentation\cite{meng2024segic,xu2024eviprompt,ayzenberg2024protosam,liu2023matcher,zhang2023personalize} has emerged as a promising alternative, integrated with vision foundation models\cite{oquab2023dinov2,CLIP}, particularly the Segment Anything Model (SAM) \cite{kirillov2023segment,ravi2024sam2}. SAM’s proven zero-shot generalization and promptable interface allow for high-quality mask generation from a single labeled \textit{support} image to \textit{query} images without any parameter updates, offering a fast solution for label-scarce medical scenarios.

The standard workflow for training-free one-shot segmentation methods follows a matching-based prompt sampling paradigm. This involves two core stages: \textbf{1)} a feature matching stage that computes a similarity heatmap between the query image and support foreground features, and \textbf{2)} a prompt sampling stage that applies various strategies to extract points, boxes, or masks for SAM input. For instance, PerSAM\cite{zhang2023personalize} and Matcher\cite{liu2023matcher} sample positive and negative points based on similarity extremes or bidirectional consistency; while OPSAM\cite{OPSAM} and ProtoSAM\cite{xu2024eviprompt} binarize heatmaps using fixed thresholds or regional statistics to extract candidate masks. These manual rules serve as the bridge between raw feature matching and the foundation model’s prompt interface.

However, we observe that the effectiveness of this paradigm is fundamentally limited by a uniformity bias, where current pipelines erroneously assume homogeneity across three critical dimensions of information. First, regarding regional homogeneity within support foreground, existing methods often treat all support foreground pixels as equally representative. In practice, colonoscopic images frequently contain suboptimal regions, such as those obscured by reflections or mucus, that introduce misleading features if not explicitly filtered. As illustrated in the top of Fig. \ref{fig:motivation}(a), matching features from different regions yields heatmaps of varying quality; specifically, degraded regions (e.g., region $d$) trigger extensive false-positive noise. Second, regarding regional homogeneity across foreground and background, many methods neglect the support background as a distinct information layer, failing to leverage it as an explicit contrastive reference. As shown in the bottom of Fig. \ref{fig:motivation}(a), utilizing background features as negative anchors is essential for suppressing noise and distinguishing polyps from similar-looking intestinal folds. Finally, regarding intensity heterogeneity, the transition from heatmaps to prompts is currently governed by static rules that ignore the stochastic response intensities across diverse query images. As evidenced by the histograms in Fig. \ref{fig:motivation}(b), the optimal binarization thresholds fluctuate significantly, making rigid sampling rules inadequate for ensuring both prompt fidelity and diversity in varied clinical conditions.



To mitigate this uniformity bias and ensure high-fidelity knowledge transfer, we propose RPG-SAM, a robust framework that explicitly accounts for information heterogeneity. We first introduce Reliability-Weighted Prototype Mining (RWPM) to resolve regional heterogeneity within the support foreground and across foreground and background; it distills support features according to their intrinsic reliability while leveraging background anchors as explicit contrastive references. To address intensity heterogeneity, we develop Geometric Adaptive Selection (GAS), which replaces static sampling rules with a dynamic thresholding mechanism guided by morphological priors. Finally, a Prior-guided Iterative Refinement (PIR) loop is designed to progressively polish anatomical boundaries. By systematically tackling these heterogeneous cues, RPG-SAM maximizes the utility of sparse support information and maintains high robustness across diverse endoscopic conditions.

\section{Methodology}

\begin{figure}
\centering
\includegraphics[width=\textwidth]{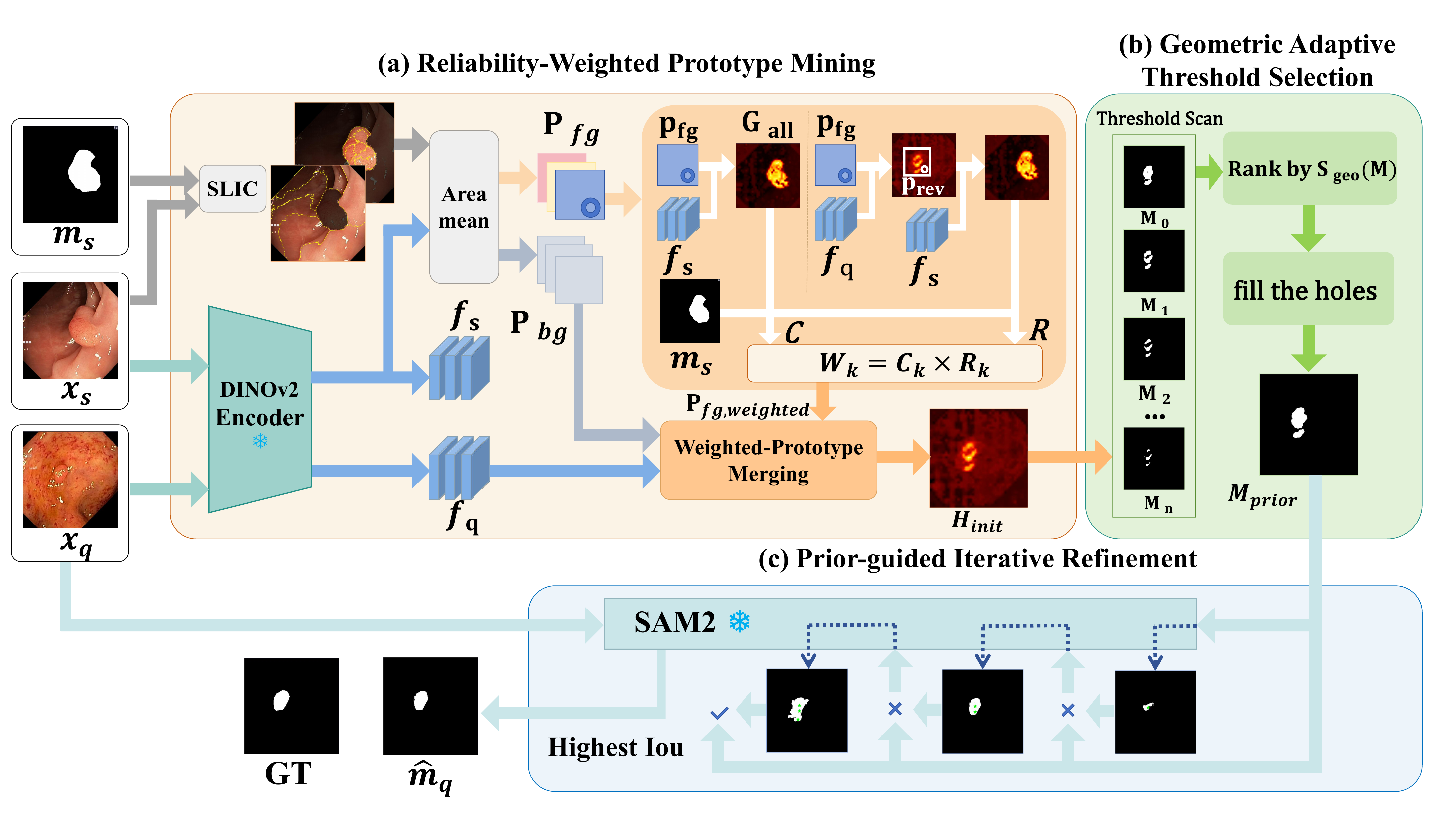}
\caption{The overview of RPG-SAM. (a)RWPM distills high-fidelity prototypes by weighting support features and suppressing background noise. (b)GAS dynamically selects optimal binarization thresholds based on morphological solidity and scale consensus. (c)PIR utilizes SAM2 to progressively polish segmentation boundaries through automated error correction.} 
\label{fig:overview}
\end{figure}

RPG-SAM is a SAM2-based training-free framework designed to segment polyps in a query image $x_q$ using a single labeled support image $x_s$ and its corresponding mask $m_s$. As illustrated in Fig. \ref{fig:overview}, the framework consists of three main components: Reliability-Weighted Prototype Mining (RWPM), Geometric Adaptive threshold Selection (GAS), and Prior-guided Iterative Refinement (PIR). We introduce the components in the following subsections.

\subsection{Reliability-Weighted Prototype Mining}
\label{sec:prototype}
Deal with the regional and contextual heterogeneity, We utilize DINOv2\cite{oquab2023dinov2} as a feature extractor to obtain deep features $f_s, f_q \in \mathbb{R}^{H'W' \times D}$ from $x_s$ and $x_q$. $x_s$ is divided into foreground and background regions according to $m_s$. To handle the complexity of endoscopic scenes, we apply the SLIC\cite{SLIC} algorithm to the support image to generate $K$ superpixel clusters. By averaging features within these clusters according to $m_s$, we obtain the foreground prototypes $\mathbf{P}_{fg} = \{p_{fg}^k\}_{k=1}^K$ and background prototypes $\mathbf{P}_{bg} = \{p_{bg}^k\}_{k=1}^K$.

To distinguish the reliability of $\mathbf{P}_{fg}$, we introduce a module that evaluates each $\mathbf{P}_{fg}^k$ based on two complementary metrics: \textbf{Intrinsic Reliability} (Contrast Factor) and \textbf{Query-specific Relevance} (Reverse Purity Factor).
\textbf{First}, we compute the Contrast Factor $\textbf{C}_k$ to evaluate discriminative power of a prototype within the support image. We compute the similarity between $p_{fg}^k$ and all support features $f_s$, put the similarity scores $\mathbf{G}_{all}$ into foreground sets $\mathbf{G}_{fg}$ or background sets $\mathbf{G}_{bg}$ according to the mask $m_s$. 
The contrast is defined as the standardized difference between these sets:
\begin{equation}
\textbf{C}_k = \text{ReLU}\left( \frac{\text{mean}(\mathbf{G}_{fg}) - \text{mean}(\mathbf{G}_{bg})}{\text{std}(\mathbf{G}_{all})} \right).
\end{equation}
Prototypes that fail to distinguish themselves from the support background (e.g., those corrupted by specular reflections) receive a lower $\mathbf{C}_k$. 
\textbf{Second}, we use a Reverse Purity Factor $\textbf{R}_k$ to verify the cross-image matching stability of $\mathbf{P}_{fg}^k$. We identify its top-$n$ matches in the query feature map $f_q$, average them to form a proxy $p_{rev}^k$, and project this proxy back onto $f_s$. The purity $p$ is the proportion of top-$n$ reverse matches falling within $m_s$. To account for varying lesion scales, we normalize this purity against the foreground area ratio $p_0$:
\begin{equation}
\textbf{R}_k = \text{ReLU}\left( \frac{p - p_0}{1 - p_0} \right).
\end{equation}
We dynamically weight each prototype's contribution based on its relevance to the current query, where $\textbf{W}_k = \textbf{C}_k \cdot \textbf{R}_k$.

Finally, we generate an initial heatmap $\mathbf{H}_{init}$ by aggregating the weighted foreground similarity maps. Unlike methods that focus solely on foreground cues, we explicitly leverage $\mathbf{P}_{bg}$ as negative anchors to suppress false-positive activations in heatmap:
\begin{equation}
\mathbf{H}_{init} = \text{Softmax}\left( K_{fg} \sum_{k=1}^{K_{fg}} W_k \cdot (f_q \cdot {p_{fg}^k}^\top) - \sum_{k=1}^{K_{bg}} (f_q \cdot {p_{bg}^k}^\top) \right).
\end{equation}
To enhance spatial consistency and reduce noise, we apply $N$ iterations of a self-diffusion mechanism\cite{OPSAM}, updating $\mathbf{H}_{init}$ by $\mathbf{H}_{init} = \mathbf{H}_{init} \cdot (f_q \cdot f_q\top)^k $.

\subsection{Geometric-Prior-based threshold Selection}
\label{sec:geo}
To mitigate the intrinsic stochasticity of response intensities across different query images, we develop Geometric Adaptive threshold Selection (GAS). Instead of using a fixed binarization threshold, GAS dynamically evaluates a pool of candidate masks to identify the one that best conforms to the expected geometric properties of a polyp.

We first generate a set of candidate binary masks $\{\mathbf{M}_{\tau}\}$ by thresholding the initial heatmap $\mathbf{H}_{init}$ across a range of potential levels $\tau \in [\tau_{min}, \tau_{max}]$. For each $\mathbf{M}_{\tau}$, we perform connected component analysis and retain only those components whose area is at least 20\% of the largest component in that mask. To ensure morphological completeness, internal holes within these valid regions are filled, resulting in refined candidate masks.
Then, each refined candidate $\mathbf{M}$ is evaluated by a geometric score $\mathbf{S}_{geo}$, which balances morphological solidity and scale consensus:
$$\mathbf{S}_{geo}(\mathbf{M}) = \underbrace{\sum_{i} \frac{|\mathbf{C}_i|}{|\mathbf{M}|} \frac{|\mathbf{C}_i|}{|\text{Hull}(\mathbf{C}_i)|}}_{\text{Weighted Solidity}} \cdot  \underbrace{\min \left(1, \frac{|\mathbf{M}|}{\mathbf{A}_{ref}} \right) }_{\text{Scale Consensus}},$$
where $\mathbf{C}_i$ denotes the $i$-th valid component of $\mathbf{M}$, and $|\cdot|$ represents the area. The weighted solidity computes the area-weighted average compactness of the components, favoring regions with regular, convex-like anatomical shapes. 
To prevent small, fragmented noise from obtaining artificially high solidity scores, we introduce Scale Consensus. This term downweights candidates that are significantly smaller than a reference area $\mathbf{A}_{ref}$, which represents the expected median scale of polyps. The candidate mask with the highest $\mathbf{S}_{geo}$ is selected as the optimal prior mask $\mathbf{M}_{prior}$ to generate sparse prompts for SAM2.

\subsection{Prior-guided Iterative SAM2 Refinement}
\label{sec:refinement}
To fully automate the segmentation process without manual intervention, we implement an iterative refinement strategy that utilizes the geometric prior $\mathbf{M}_{prior}$ to guide SAM2. This process progressively optimizes the segmentation mask $\mathbf{M}_t$ by using the structural consistency of the prior as a reference for error correction. In each iteration, the current mask is evaluated against $\mathbf{M}_{prior}$ based on Coverage ($Cov$) and Intersection-over-Union ($IoU$), allowing the framework to utilize SAM2’s edge-refinement capabilities to polish anatomical boundaries.

The refinement logic follows a hierarchical approach by first addressing insufficient lesion coverage ($Cov < \tau_{cov}$). When such errors are detected, the system identifies the false-negative region $\mathbf{R}_{FN} = \mathbf{M}_{prior} \cap \neg \mathbf{M}_t$ and samples its geometric center via the Euclidean Distance Transform (EDT) as a positive prompt to expand the mask. Conversely, if coverage is adequate but the $IoU$ remains suboptimal, the focus shifts to suppressing background noise by identifying the false-positive region $\mathbf{R}_{FP} = \mathbf{M}_t \cap \neg \mathbf{M}_{prior}$ and inserting a negative prompt. This iterative correction continues until the stopping criteria $(Cov_{t} \geq \tau_{cov}) \land (IoU_{t} \geq \tau_{iou})$ are satisfied or the maximum iteration count $T_{max}$ is reached. To ensure robustness, the mask with the highest $IoU$ relative to the prior throughout the iteration history is selected as the final query prediction $\hat{m}_q$.

\section{Experiment and Evaluations}
\subsection{Experiment Settings}
\textbf{Datasets and Metrics.} 
We validate the effectiveness of RPG-SAM on four widely used public polyp segmentation datasets: \textbf{Kvasir}\cite{datasetKvasir}, \textbf{CVC-ClinicDB}\cite{datasetCVC-ClinicDB}, \textbf{CVC-ColonDB}\cite{datasetCVC-ColonDB} and three-center \textbf{PolypGen}\cite{datasetPolypGEN}. 
Following OPSAM\cite{OPSAM}, we randomly select one image as the support sample and keep it fixed for all methods to ensure fair comparison. We employ the mean Dice coefficient (mDice) and mean Intersection over Union (mIoU) as the primary evaluation metrics.

\noindent \textbf{Implementation Details}
Our framework is implemented in PyTorch. We use DINOv2\cite{oquab2023dinov2} with a ViT-L/14 as the feature encoder with an input resolution of $560 \times 560$, and SAM2\cite{ravi2024sam2} as the mask generator with an input size of $1024 \times 1024$. 
The hyperparameter settings are as follows:
(1) For \textbf{RWPM}, the SLIC parameters are set to $K=10$ and $m=20$.
(2) For \textbf{GAS}, the confidence scanning range is set to $[0.4, 0.7]$ with a stride of 0.05.
(3) For the \textbf{PIR}, the stopping thresholds are set to $\tau_{cov}=0.9$ and $\tau_{iou}=0.8$, with a maximum iteration count of $T_{max}=5$.
All inference processes are conducted on a single NVIDIA RTX 3090 GPU without any model training or fine-tuning.


\begin{table*}[t]
\centering
\caption{Compared with SOTA methods. The best results are highlighted in \textbf{bold}, and the second-best results are \underline{underlined}.}
\label{tab:comparison}
\resizebox{\textwidth}{!}{%
\begin{tabular}{l|cc|cc|cc|cc|cc|cc}
\hline
\multirow{3}{*}{\textbf{Method}} & \multicolumn{2}{c|}{\multirow{2}{*}{\textbf{Kvasir}}} & \multicolumn{6}{c|}{\textbf{PolypGen}} & \multicolumn{2}{c|}{\multirow{2}{*}{\textbf{CVC-ClinicDB}}} & \multicolumn{2}{c}{\multirow{2}{*}{\textbf{CVC-ColonDB}}} \\ \cline{4-9}
 & \multicolumn{2}{c|}{} & \multicolumn{2}{c|}{\textbf{Center 1}} & \multicolumn{2}{c|}{\textbf{Center 2}} & \multicolumn{2}{c|}{\textbf{Center 3}} & \multicolumn{2}{c|}{} & \multicolumn{2}{c}{} \\ \cline{2-13}
 & IoU & Dice & IoU & Dice & IoU & Dice & IoU & Dice & IoU & Dice & IoU & Dice \\ \hline
SEGIC\cite{meng2024segic} & 22.23 & 32.81 & 13.36 & 20.73 & 15.47 & 22.79 & 15.73 & 23.77 & 17.87 & 26.98 & 12.00 & 18.25 \\
PerSAM\cite{zhang2023personalize} & 26.01 & 35.38 & 26.15 & 31.47 & 12.97 & 18.96 & 10.87 & 17.74 & 13.29 & 20.10 & 10.31 & 16.18 \\
OPSAM\cite{OPSAM} & 63.95 & 71.43 & \underline{57.16} & 62.53 & 59.29 & 65.33 & \underline{74.91} & \underline{81.67} & 58.75 & 65.60 & \underline{56.44} & 62.86 \\
Matcher\cite{liu2023matcher} & 71.71 & 78.91 & 56.25 & \underline{63.06} & \underline{60.79} & \underline{66.30} & 65.49 & 71.42 & 63.39 & 70.20 & 52.01 & 58.42 \\
ProtoSAM\cite{ayzenberg2024protosam} & \underline{73.09} & \underline{81.54} & 51.14 & 58.63 & 55.73 & 63.03 & 69.79 & 76.96 & \underline{68.99} & \underline{76.83} & 55.92 & \underline{63.45} \\ \hline
\textbf{RPG-SAM (Ours)} & \textbf{78.65} & \textbf{85.65} & \textbf{59.45} & \textbf{64.78} & \textbf{61.04} & \textbf{67.14} & \textbf{76.75} & \textbf{83.56} & \textbf{70.18} & \textbf{77.76} & \textbf{61.16} & \textbf{68.77} \\ \hline
\end{tabular}%
}
\end{table*}

\subsection{Comparison with Baseline Models}
We compare RPG-SAM against SOTA one-shot methods, including \textbf{SEGIC}\cite{meng2024segic}, \textbf{PerSAM}\cite{zhang2023personalize}, \textbf{Matcher}\cite{liu2023matcher}, \textbf{OPSAM}\cite{OPSAM}, and \textbf{ProtoSAM}\cite{ayzenberg2024protosam}. For fairness, all methods utilize identical support-query pairs and backbone settings. Results are detailed in Table \ref{tab:comparison}. RPG-SAM consistently outperforms existing paradigms. Notably, on Kvasir, it achieves 78.65\% mIoU and 85.65\% mDice, surpassing ProtoSAM by 5.56\% and 4.11\%, respectively. This improvement stems from RWPM, which discriminatively prioritizes reliable support features and suppresses toxic features like specular reflections, unlike methods assuming uniform representation. On the multi-center PolypGen dataset, RPG-SAM demonstrates superior robustness. While domain shifts often trigger false-positive activations in other models, our Background Suppression and GAS stabilize response intensities across centers. This confirms RPG-SAM’s clinical reliability beyond single-center data. Unlike fixed-threshold methods (e.g., OPSAM), the GAS module adaptively recalibrates for each query image. This prevents prompt contamination in low-contrast cases, ensuring the iterative refinement loop operates on high-quality, diverse candidates to polish anatomical boundaries.

\begin{figure}[t]
\centering
\includegraphics[width=\textwidth]{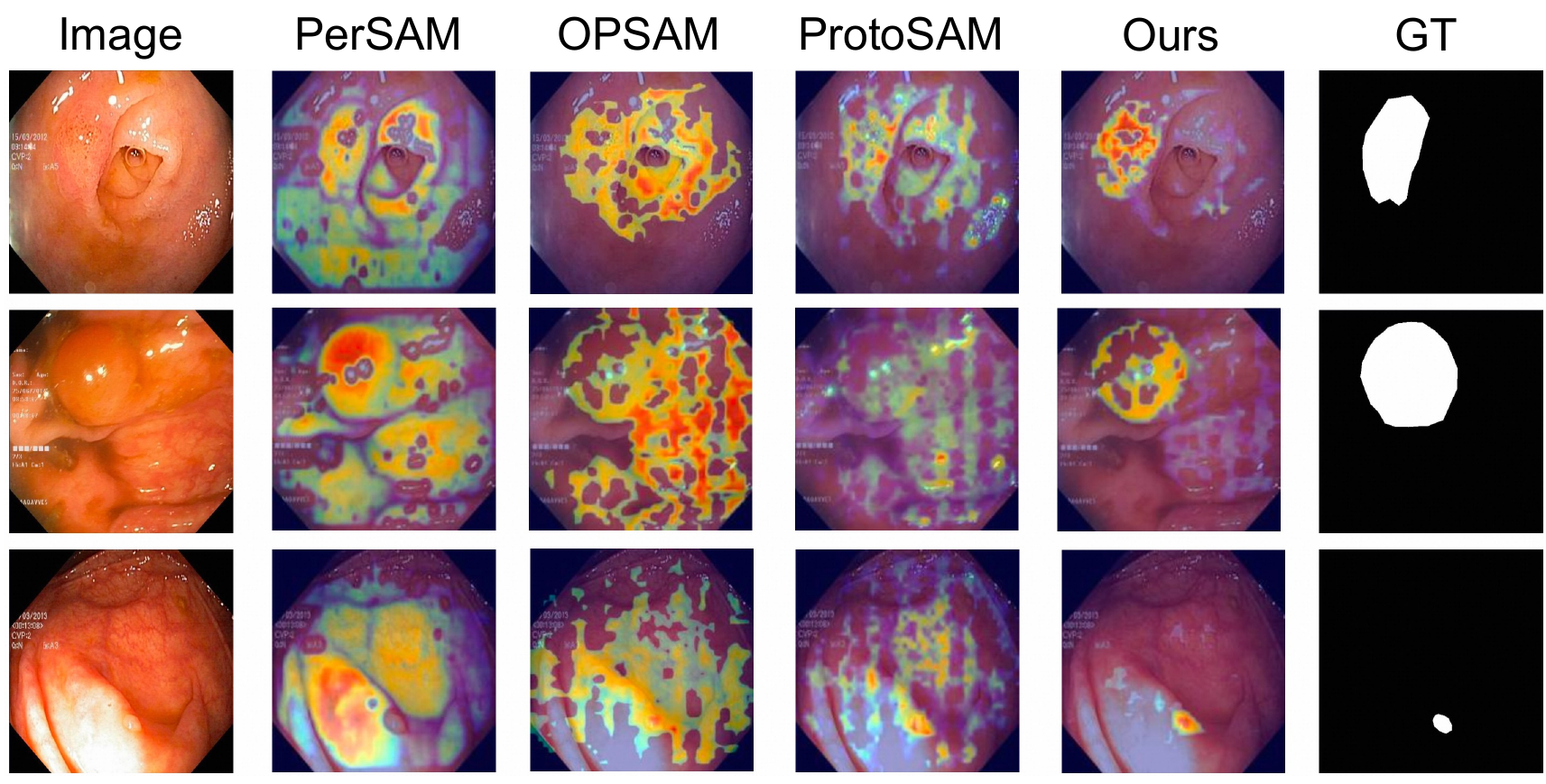}
\caption{Visualization of heatmaps on challenging query samples, including cases with small ground-truth regions or the presence of visually similar distractors.} 
\end{figure}

\subsection{Ablation Studies}
We evaluate the components and hyperparameter sensitivity of RPG-SAM on the Kvasir dataset. As shown in Table \ref{tab:ablation_module}, each module contributes significantly to the final performance. The baseline without background suppression performs poorly due to unconstrained feature transfer. Explicitly introducing Background Suppression (BG Supp.) yields a 3.78\% gain in mDice, proving its necessity in penalizing false-positive activations. RWPM further improves spatial granularity, relieving the uniform representation fallacy. The GAS module outperforms fixed thresholds ($\tau=0.7$, optimal) by 2.59\% mDice, effectively mitigating response intensity stochasticity. Finally, PIR further improving the results.

\begin{table}[t]
\centering
\caption{Incremental ablation on Kvasir dataset. We demonstrate RPG-SAM benefits of Background Suppression (BG Supp.), RWPM, GAS, and PIR.}
\label{tab:ablation_module}
\resizebox{0.9\linewidth}{!}{
\begin{tabular}{cccc|ccc}
\hline
\textbf{BG Supp.} & \textbf{RWPM} & \textbf{GAS} & \textbf{PIR} & \textbf{mIoU (\%)} & \textbf{mDice (\%)} & \textbf{AUC-PR} \\ \hline
 - & - & - & - & 63.68 & 74.94 & 0.8743 \\
 \checkmark & - & - & - & 67.89 & 78.72 & 0.8921 \\
 \checkmark & \checkmark & - & - & 71.27 & 81.05 & 0.9068 \\ 
 \checkmark & \checkmark & \checkmark & - & 75.25 & 83.64 & 0.9068 \\ 
 \textbf{\checkmark} & \textbf{\checkmark} & \textbf{\checkmark} & \textbf{\checkmark} & \textbf{78.65} & \textbf{85.65} & \textbf{0.9068} \\ \hline
\end{tabular}
} 
\end{table}

We further analyze the robustness of hyperparameters in Table \ref{tab:ablation_params}. For GAS, the scanning range $[0.4, 0.7]$ provides the optimal candidate pool; lower ranges introduce excessive noise, while higher ranges limit diversity. For RWPM, SLIC parameters $K=10$ and $m=20$ strike the best balance between spatial accuracy and regularization. Finally, the framework exhibits high stability regarding PIR thresholds, with $\tau_{cov}=0.9$ and $\tau_{iou}=0.8$ achieving the optimal balance between recall and precision.

\begin{table}[t]
\centering
\caption{Ablation study on hyper-parameters for SLIC, GAS, and PIR modules. Params in \textbf{bold} are used in RPG-SAM.} 
\label{tab:ablation_params}
\resizebox{0.9\linewidth}{!}{
\begin{tabular}{cc|cc|cc|cc|cc|cc}
\hline
\multicolumn{4}{c|}{\textbf{SLIC params}} & \multicolumn{4}{c|}{\textbf{PIR params}} & \multicolumn{4}{c}{\textbf{GAS params}} \\ \hline
\textbf{K} & \textbf{m} & \textbf{mIoU} & \textbf{mDice} & \textbf{$\mathbf{\tau_{cov}}$} & \textbf{$\mathbf{\tau_{iou}}$} & \textbf{mIoU} & \textbf{mDice} & \multicolumn{2}{c|}{\textbf{Range}} & \textbf{mIoU} & \textbf{mDice} \\ \hline
5  & 5  & 77.34 & 84.52 & 0.8  & 0.7 & 77.98 & 85.20 & \multicolumn{2}{c|}{0.3-0.6} & 75.05 & 82.57 \\
10 & 5  & 78.28 & 85.27 & 0.8  & 0.8 & 78.11 & 85.32 & \multicolumn{2}{c|}{0.4-0.6} & 77.15 & 84.48 \\
\textbf{10} & \textbf{20} & 78.65 & 85.65 & 0.8  & 0.9 & 77.79 & 85.12 & \multicolumn{2}{c|}{\textbf{0.4}-\textbf{0.7}} & 78.65 & 85.65 \\
10 & 50 & 78.34 & 85.41 & \textbf{0.9}  & \textbf{0.8} & 78.65 & 85.65 & \multicolumn{2}{c|}{0.5-0.7} & 77.26 & 84.57 \\
20 & 5  & 76.08 & 83.21 & 0.95 & 0.8 & 78.66 & 85.60 & \multicolumn{2}{c|}{0.5-0.8} & 77.12 & 84.41 \\ \hline
\end{tabular}
} 
\end{table}

\section{Conclusion}
We proposed RPG-SAM, a training-free framework addressing noise and intensity stochasticity in one-shot polyp segmentation. RWPM resolves the uniform representation fallacy through reliability-driven feature distillation and background suppression, while GAS replaces rigid thresholds with adaptive binarization based on geometric priors. Extensive benchmarks, including multi-center cohorts, confirm the effectiveness of RPG-SAM in terms of generalization and enhanced performance. Our approach provides a robust, scalable alternative to data-intensive models in label-scarce clinical settings. Future research will extend this framework to exploit temporal consistency in endoscopic videos.

%
%
%
%
\bibliographystyle{splncs04}
\bibliography{refs}
\end{document}